# Machine Learning Advances aiding Recognition and Classification of Indian Monuments and Landmarks

Aditya Jyoti Paul [1,2]
aditya_jyoti@carlresearch.org

Smaranjit Ghose [1,2]
smaranjitghose@protonmail.com

Kanishka Aggarwal [1,4]
kanishkaaggarwal.cse2@bvp.edu.in

Niketha Nethaji [1,2]
nikenethaji@gmail.com

Shivam Pal [1,3]
shivam_pal@srmuniv.edu.in

Arnab Dutta Purkayastha [1,2]
arnabduttapurkayastha@gmail.com

1. Cognitive Applications Research Lab, India.
2. Department of Computer Science and Engineering, SRM Institute of Science and Technology, Chennai, India.
3. Department of Electronics and Communication Engineering, SRM Institute of Science and Technology, Chennai, India.
4. Department of Computer Science and Engineering, Bharati Vidyapeeth's College of Engineering, New Delhi, India.

*Abstract*— Tourism in India plays a quintessential role in the country's economy with an estimated 9.2% GDP share for the year 2018. With a yearly growth rate of 6.2%, the industry holds a huge potential for being the primary driver of the economy as observed in the nations of the Middle East like the United Arab Emirates. The historical and cultural diversity exhibited throughout the geography of the nation is a unique spectacle for people around the world and therefore serves to attract tourists in tens of millions in number every year. Traditionally, tour guides or academic professionals who study these heritage monuments were responsible for providing information to the visitors regarding their architectural and historical significance. However, unfortunately this system has several caveats when considered on a large scale such as unavailability of sufficient trained people, lack of accurate information, failure to convey the richness of details in an attractive format etc. Recently, machine learning approaches revolving around the usage of monument pictures have been shown to be useful for rudimentary analysis of heritage sights. This paper serves as a survey of the research endeavors undertaken in this direction which would eventually provide insights for building an automated decision system that could be utilized to make the experience of tourism in India more modernized for visitors.

## I. Introduction

India possesses a rich and diverse cultural and historical heritage, the maintenance and conservation of which is of pivotal importance in today's fast-paced world. Archaeologists and historians have put in a lot of time and effort studying the different monuments and architectural styles by visiting the sites and making first-hand observations. Building upon their work and making a part of that process more streamlined and scalable, Computer Vision techniques have now made a foray and are also being used to study the monuments, the applications of which involve classification of monuments, segmentation of individual architectural styles and much more.

Monument classification can be broadly described as the task of identifying and classifying images of monuments into sub-categories based on their architectural style. Monument recognition and classification comes under the broader domain of landmark recognition. Though landmark recognition is a well-researched area in computer vision, monument recognition still remains challenging due to various factors such as the lack of sufficient annotated datasets of monuments in non-English speaking regions, subtle variations in architectural styles of monuments, image samples being varied in perspective, resolution, lighting, scale and viewpoint. These significant challenges make monument recognition in a diverse region such as India significantly non-trivial. Automatic recognition of monuments provides wide-ranging applications in multiple domains, including but not limited to archaeological, historical research, conservation, tourist attraction and education.

This paper will present a survey on previous work done in the field of monument recognition and classification. This research. The rest of the paper is arranged as follows, Section II discusses the Preliminaries going over the architectural foundations necessary to dive deep into the paper, Section III provides a thorough survey of methods used for feature extraction, recognition, information retrieval and classification, Section IV discusses the authors' ongoing research and Section V serves as a conclusion and suggests some future avenues.

## II. Preliminaries

Throughout the timeline of Indian history, the architectural styles have evolved and been influenced from time to time by factors like the patronage of the ruling dynasties, foreign invasions and shift in spiritual and religious trends like the Bhakti movement [1]. Therefore, having a unanimous categorization of the heritage sites in the nation is not an easily achievable task. There have been various classifications based upon religious or communal ties of the monuments like Sikh architecture, Mughal architecture etc. On the other hand, temples itself had distinguishable categories like Dravidian and Nagara architecture. In this section, some of the common categories are discussed.

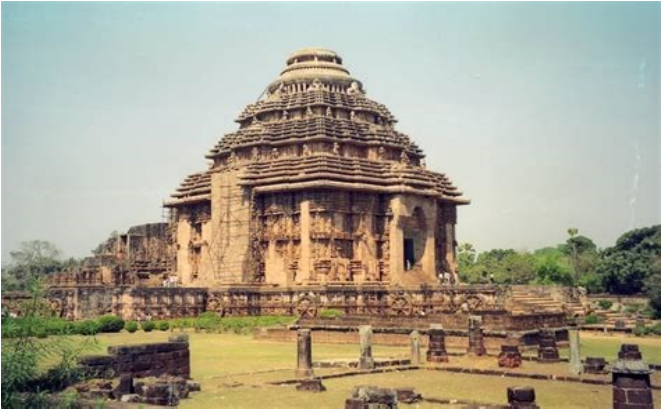

Fig. 1. Sun Temple, Konark.

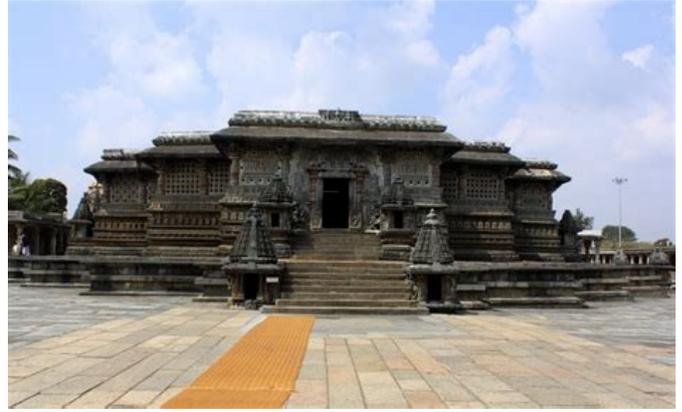

Fig. 3. Chennakeshava Temple, Belur

*1. Nagara Architecture:* This style was common among the temples initially built in the northern, central and western regions of the Indian subcontinent. These temples have several characteristic features. There are multiple shikharas above the sanctum which are the most prominent characteristic of the temple. The sanctums can be either single story or multi-story with a curvilinear central tower. The pedestals are at a higher level than the surrounding land. Temple boundaries are not given much importance with respect to the details. The image given in Fig. 1 is an example of Nagara architectural style.

*2. Dravidian Architecture:* This style was common among the temples initially built in the northern, central and western regions of the Indian subcontinent. These temples have several characteristic features. There are multiple shikharas above the sanctum which are the most prominent characteristic of the temple. The sanctums can be either single story or multi-story with a curvilinear central tower. The pedestals are at a higher level than the surrounding land. Temple boundaries are not given much importance with respect to the details. The image given in Fig. 2 is an example of Dravidian architectural style.

*3. Vesara Architecture:* This architectural style is an amalgamation of the Dravidian and Nagara styles which gained prominence during the era of the Chalukya dynasty. The superstructures like the shikharas are influenced by the Nagara style while the mandapas are influenced by the Dravidian style. Stellate arrangement is common among the walls of the temple with intricate ornamentation representing both the styles. Adhishthana is the high platform built for raising the temples above the surrounding ground. These temples are majorly found in the Deccan region. The image given in Fig. 3 is an example of Vesara architectural style

*4. Mughal Architecture:* This style rose to prominence during the era of Mughal invasion of India. It was initially an incorporation of features of the Persian architectures on the existing Indian temple architectures but slowly evolved to be a standalone style. Some of these monuments have gained a reputation for being a unique spectacle in the world like the famous Taj Mahal. A common characteristic of these monuments is having four minarets on each side of the building. Spheroid domes, palace halls built on large pillars and huge entrances were among the other common features. The majority of the monuments including mausoleums, palaces and mosques were built using red or white sandstone. Although monuments previously built for Muslim Indian rules were existent under the umbrella of Indo-Islamic style, the extravagant decor of the Mughal architecture set it apart. The image given in Fig. 4 is an example of Mughal architectural style.

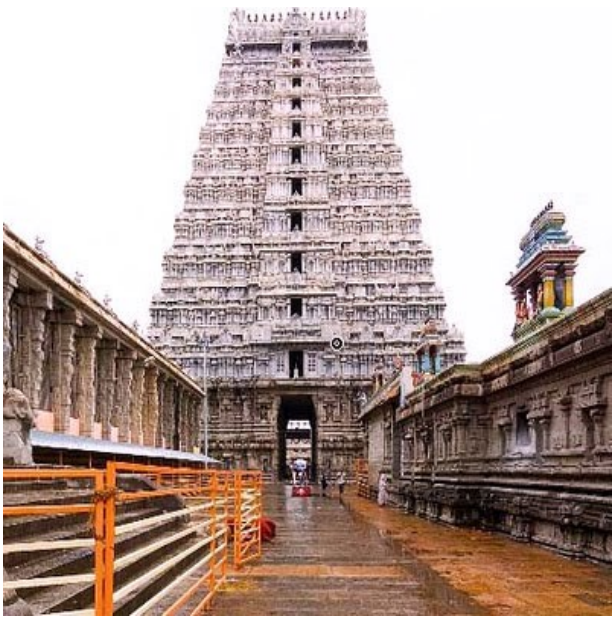

Fig. 2. Arunachalesvara Temple, Tiruvannamalai

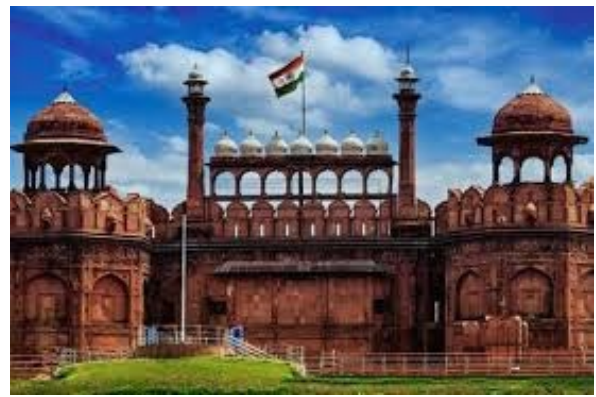

Fig. 4. Red Fort, New Delhi

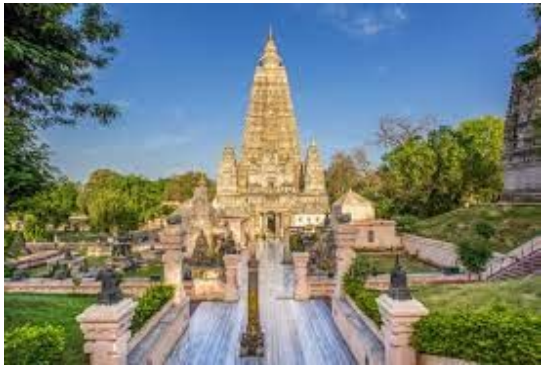

Fig. 5. Bodh Gaya, Bihar

*5. Buddhist Architecture:* This style is exclusively found among the religious buildings of Buddhism. It first flourished in the Indian subcontinent under the patronage of the emperor Ashoka and then expanded across South East Asia with variances in the style according to the region. The buildings of this style can be primarily categorized into three types: the worship halls called the Chaitya, the monasteries known as Vihara and the hemispherical mounds known as Stupa. The majority of these monuments have walls engraved with the teachings of Buddha, gates with semicircular arches, a topmost spike representing five fundamental elements of existence and a boundary made of lines of columns. The image given in Fig. 5 is an example of Buddhist architectural style.

*6. Sikh Architecture:* This style is exclusively found among the monuments which flourished during the era of the Sikh empire. Although majority of the buildings exhibiting this style were religious buildings for the Sikhs, the style found its way into non-religious buildings like bangas or palaces, colleges, etc. with the passage of time. The presence of four entrances which is symbolic of being able to enter without any obstruction is commonly found in several of these monuments especially the gurudwaras. Elliptical or ogee arches, bulbous domes covered capped with brass or copper roofs and miniature shrines are amongst other typical features. Although, a building of this style especially gurudwaras are found in various regions of the country, the heritage sites are situated in the Punjab region of the Indian subcontinent. The image given in Fig. 6 is an example of Sikh architectural style.

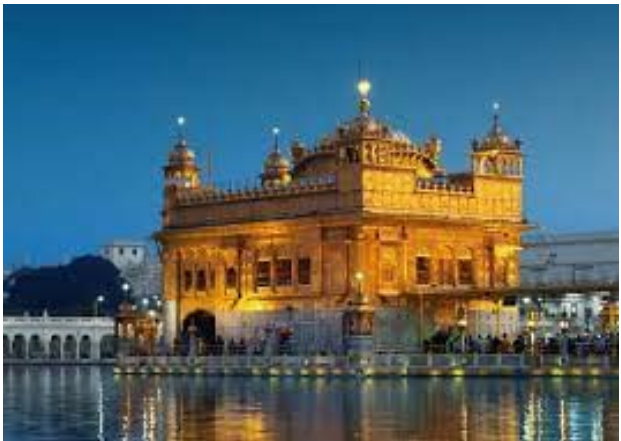

Fig. 6. Harmandir Sahib, Amritsar

### III. LITERATURE REVIEW

Researchers have been working on landmark classification, which is a superset of monument classification since the past few decades, using various techniques that can be global feature based or local feature based. Global features like textures, edges and colors are rudimentary and least resource-intensive of the two. Linde et al. [2] demonstrated the superiority of higher order composed field histograms leveraging efficient computation on sparse matrices. Ge et al. [3] showed a covariance descriptor based technique using a Support Vector Machine (SVM) for the combined classifiers and voting strategy. Visual context was utilized for place recognition and categorization by Torralba et al. [4] and the authors demonstrated how information about the scene could be exploited for object detection by contextual priming. In 2020, a novel method involving an ensemble of subcenter ArcFace models [5] with dynamic margins, using only global features, won the Google Landmark Recognition challenge on the GLDv2 dataset [6].

However, since global features lack granularity and cannot focus on Regions of Interest (ROIs), they are usually used in conjunction with Local features for monument detection or the object detection class of problems in general. Local features are focused on Points of Interest (POI) or Regions of Interest (ROI) and are robust to partial occlusion, illumination variance and changes in viewpoint. Standard approaches include Scale-invariant Feature Transform (SIFT) [7] and affine-invariant features [8]. These approaches commonly use a Bag-of-Words (BoW) model [9,10] on the clustered visual words, representing local features. Many such methods have been presented which include using probability density response map for local patches likelihood [11], patch saliency estimation using contextual knowledge [12], calculating the importance of patches through non-parametric density estimation [13], spatial pyramid kernel-based BoW approaches (SPK-BoW) [14,15] and scalable vocabulary trees [16].

Content-based Image Retrieval (CBIR) is another research area in the applications of Computer Vision in monuments and landmarks recognition. Desai et al. [17] discussed a CBIR system that classifies Indian monuments using various local features. Morphological operations are used for shape features extraction and Gray Level Co-occurrence Matrix (GLCM) is used for texture features. The paper considered 500 images belonging to 5 classes: Mosques, Churches, Hampi Temples, Kerala Temples and Southern temples having a trapezoid shape, and reported a retrieval accuracy of 78-90% for the top 10 images per class. The experiments performed much better on the average weighted precision metric as compared to Sobel [18] and Canny [19] detectors. This work seemed to build upon the similar findings by [20] who built a classifier for different architectural styles of facade windows belonging to Gothic, Romanesque, Gothic and Baroque periods. Similarly, [21] also proposed a CBIR system using morphological gradients and GLCM. Fuxiang Lu et al. [22] proposed an Improved Local Binary Pattern (ILBP) that is more efficient compared to basic LBP methods in terms of discovery of more basic primitives like straight lines, T-junctions and cross-intersections. [23]

discussed a CBIR technique using Ant Colony Optimization (ACO). [24] proposed an efficient CBIR system which takes into account color features from RGB histogram in addition to shape and texture features from mathematical morphology and ILBP respectively.

An investigation on the utility of the characteristic information derived out of low-level feature configurations for the classification of monuments into architectural style using a European monument image dataset was done by [25]. Rischka et al. [26] proposed a landmark recognition system using a hierarchical K-Means clustering system. In [27], Kim et al. prototyped a classification and indexation scheme from a large image repository using local features, user tags and GPS features. The authors in [28] proposed a monuments classification system with 9 categories having labels like Taj Mahal, Qutub Minar, Red Fort using various hand-crafted features along with Bag-of-Words descriptors. They also compared various existing approaches on their dataset.

Sharma et al. [29] curated an Indian monuments dataset and proposed two approaches to classify them, one using Radon Barcodes achieving an accuracy of 76%, and another CNN based approach with an accuracy if 82%. The Radon transform was inspired by the Radon transform normalization discussed in [30]. Yi et al. [31] discusses automated house style classification using CNNs. [32] proposed an automated architectural style recognition system from façade images.

In 2015, Gupta and Chaudhury [33] had demonstrated the use of a 7 layer DCNN and ontology priors for training a monument classification model on a dataset made up of 1500 images of forts, tombs and mosques which was a drastic improvement over a simple logistic classifier. Bhatt and Patalia used a genetic algorithm [34] based approach for building a Indian monument classification system. They considered using 100 images with 25 image each of Golden Temple, India Gate, Qutub Minar and Taj Mahal for training the model. To ensure the robustness of the model, multiple views like right side view, front view, left side view and far view were considered. They achieved an accuracy of 92.75% when the system reached the 100th generation after 7 hours of training.

Gada et al. [35] used an Inception Net [36] based Deep Convolutional Neural Network (DCNN) in 2017 study to classify the 12 most famous monuments found across the Golden Quadrilateral. The researchers observed a training accuracy of 99.4% after training the model for 4000 epochs with their transfer learning [37] approach. In the same year, another group of researchers [38] used a dataset comprising 100 monuments with 50 images per class to compare and contrast the efficacy of Support Vector Machine (SVM) [39], Random Forest [40], K-Nearest Neighbour (KNN) and a simple DCNN as classifiers for the task of monument identification. An interesting highlight of their work was the use of Graph Based Visual Saliency (GBVS) [41] to improve the model accuracy. This seemed to corroborate the claims of a similar 2013 [42] work which involved the classification of monuments found in the areas of Crete, Heraklion and Greece by using Speeded Up Robust Features (SURF) [43] and Scale Invariant Feature Transform (SIFT). Previously, [44] had conducted a study that revealed the performance boost on usage of local features for the automatic identification of landmarks as compared to global features.

Bergado et al. [45] had used airborne images to train a DCNN for the classification of buildings in urban areas. This could serve as precursor to training models using aerial images obtained from drones to provide real time inference on monuments or heritages sites. The study corroborated the notion of the robustness of a DCNN for the extraction of spatio-contextual features.

Shukhla et al. [46] used a DCNN for extraction of image features and SVM and KNN as classifiers for building a monument recognition model using a dataset comprising 6102 images of heritage sites in India classified into Indo-Islamic, Cave, Colonial, Temple and Rock-Cut architecture styles. The researchers also built another classifier using the same techniques to categorize the temple related images in their dataset into Nagara, Vesara and Dravidian styles. Additionally, they used the same dataset and built a third model to classify the images according to the building. The SVM based classifier marginally outperformed the KNN based on the images involved natural landscapes.

A similar report [47] was produced by three researchers from the University of Massachusetts, Amherst whose aim was to classify images of monuments into Sikh, British, Maratha, Indo-Islamic and Ancient styles. Their experiments reported the highest accuracy for the usage Oriented FAST and Rotated BRIEF(ORB) [48] features using descriptor-wise classification with KNN.

A thorough study regarding the application of KNN for content based image classification for the task of monument classification was previously conducted by [49] where the PISA dataset [50] composed of 1,227 photos of 12 cultural sites located in and around Pisa were used. The local feature based classifier was found out to be best in overall performance with respect to the various metrics used.

Jose et.al [51] compared the usage of ResNet and Inception-ResNet-v2 based DCNNs for optimizing the task of monument identification. They used a dataset of 10000 images divided into 10 classes. Without fine-tuning the ResNet [52] based model produced the highest accuracy whereas after fine tuning the Inception-ResNet-v2 [53] based model gave a higher accuracy. The work served to provide a basis for the consideration of deep learning based techniques for the task of monument identification.

Furthermore, a 2018 study [54] threw light on the usage of both Convolutional as well Recurrent Neural Networks(RNN) [55] for the task of cultural heritage classification related to the Indonesian space. While DCNN were employed for the classification of audio, image and videos, RNN were used to do the same for text related to task. An accuracy of 92% by the RNN for text classification of 100 text files, and 76% and 57% each by the CNN on images and audio respectively were obtained. In the past, [56] demonstrated the usage of Unmanned Aerial Vehicles (UAVs) to capture high resolution imagery to generate a 3D reconstruction of the heritage sites, a case study of which was presented for the Curium archaeological site in Cyprus. Another group of researchers [57] also demonstrated the 3D classification and segmentation of cultural sites using several decision tree based classifiers like random tree, random forest, fast random forest, LogitBoost [58], J48 [59] and RepTREE [60] on point cloud data. The highest accuracy was obtained on by the fast random forest algorithm which was around 69%. Classification of Landmarks using three dimensional point clouds was discussed in [61]. A 3-D reconstruction based on attention is proposed in [62], involving a k-dimensional tree or k-d tree on SIFT features.

An end-to-end Machine Learning based system would not only involve inference from visual data but also conversational agents and certain automated text domain tasks to simulate the offering of a tour guide. Ghosh et al. [63] had proposed a system to utilize ergo centric videos taken from heritage sites to produce dynamic stories. A real time system for monument recognition was proposed in [64] which provided the information in English as well as Hindi to cater to a larger audience. SURF was used for extraction of features while SVM was utilized for the task of classification. Ninawe et al. [65] showed the usage of a DCNN built using TensorFlow framework for the classification of Catherdral and Mughal monuments in India. The dataset used for training comprised 5000 images out of which images of Parish Church, Taj Mahal, Basilica and Char Minar were selected as leaf nodes. They found out that the local weight sharing in DCNN was an important factor for producing an accuracy of 80% which outperformed the previous work done by the authors of [33].

In [66], Chen et al. proposed a novel discriminative bag- of-visual phrase approach for mobile devices, which also incorporates the location and direction data from mobiles in the predictions, achieving a recognition accuracy of 98.3%. Zhang et al. [67] proposed a system to recognize architectural styles using hierarchical sparse coding of 'blocklets', which capture morphological features using an efficient blocklet-to-blocklet matching technique.

In [68] Jankovi demonstrated the utility of decision tree based algorithms for classification of images of heritage sites. The dataset was obtained by scrapping images from Flickr and Google and comprised 150 images equally distributed into the classes: Fresco, Architectural Sites and Monasteries. Hoeffding [69], Random Forest, J48 and Random Tree algorithms were used for the task of classification. Random Forest gave the best classification accuracy while Random Tree gave the worst classification accuracy. Hoeffding Tree had a classification accuracy of 85% which was closer to the best accuracy of 90% and it also produced the lowest mean absolute error during the training process among all other algorithms. The Weka [70] machine learning toolkit was used to perform majority of the experiments. [71] demonstrated a classification network based on MobileNet-v2 on the Indian Digital Heritage Space dataset.

An extension of the work by the same author was done in [68] where algorithms like Multi-Layer Perceptron(MLP) [72], ForestPA [73], AODE [74] and RSeslibKnn [75] were used for the task of classification of images of cultural sites. The dataset used in these experiments comprised 4000 images categorized into 5 classes of Alar, Dome, Column, Vault and Gargoyle. AOD and MLP gave the best scores for multiple performance metrics like F1, Recall, ROC area, Kappa and Precision. The lowest classification accuracy was obtained by Forest PA algorithm. Additionally, a DCNN was also trained on the same dataset to achieve the same goal of cultural site classification. Although the training of the DCNN was much computationally expensive as compared to the other algorithms, it gave the highest accuracy which was 92%. However, with attribute selection the best accuracy was obtained by the MLP algorithm.

A cross-platform machine learning based system [76] was proposed in 2019 for the classification of historical sites in the country of Iran. The researchers used a VGGNet [77] based 19 layer DCNN architecture for training the classifier. An accuracy of 95% was obtained for the training dataset. The overall proposed system aims to be decentralized and platform-independent in nature where images taken from a user's smartphone camera and the device's IP address are sent to a server for getting inference.

A 2021 study [78] compared the performance of two DCNN algorithms based on ResNet and VGGNet for the task of monument recognition. The ultimate aim of the researcher was to provide a proof-of-concept application for automated tourism in Egypt. However, due the lack of sufficient data, the study was performed using a dataset publicly available on Kaggle which comprised 1,286 images. The classes into which the images were categorized were Qutub Minar, Iron Pillar, Alai Darwaza, Alai Minar and Jamali Kamali Tomb. A 3:1 split was used for training and testing purposes. The ResNet based DCNN gave the highest accuracy of 88% on the test split while on the training split, the accuracy of both the ResNet and VGGNet based models were comparable.

A survey on landmark recognition for edge devices [79] was presented in 2009. Since then, the field has seen massive development. In recent times, with advances in quantization [80-83], these deployments have become far more efficient and lightweight and have proven useful in different computer vision applications like in [84-87], with significantly unexplored avenues and scope of improvement. Having such an application for monument recognition or architectural style classification on the edge would make it more accessible to students, researchers or even casual travelers who might not always have internet access or high computational power in all the places they travel to; this application would provide them with the right set of resources and information to better analyze the historic and cultural significance of the monuments and glean various insights from the structures which would otherwise be difficult. Such applications could also be provided as a cloud run service, which would be better suited to large batches of images, providing an architectural analysis of all the photographs by a traveler or researcher during their trip.

## IV. Our Ongoing Research

Our team at Cognitive Applications Research Lab is working on building a system that recognizes and classifies a large number of monuments throughout the geography of the Indian subcontinent. Although previous works have been able to prove the ability of DCNNs for building an automated monument recognition system, the current literature lacks the evidence of the performance of such a system when used at scale for the classification of hundreds of monuments across the nation. Our ongoing experiments aim to achieve some novel results in that direction. Furthermore, with the availability of huge amounts of data and parallel compute systems like GPUs and TPUs, there have been multiple advances in the optimization of using DCNNs and other similar neural network based algorithms for the task of image classification. We aim to apply the benefits of those outcomes to this domain of monument identification and build upon the previous demonstrations of usage of DCNNs. Lastly, to compare the efficacy of majority of the previous works, there is no single dataset which would do justice to the comparative analysis. Hence our team looks forward to provide an open-source dataset to resolve this issue which would also be used for our upcoming experiments.

## V. Conclusion

This research surveys and demonstrates the current advances in the field of landmark and monument classification and recognition using Machine Learning and Computer Vision techniques with a focus on the Indian geography. It also goes over the architectural analysis and modeling of the monuments using various analytical and reconstruction techniques, and briefly covers the authors' ongoing research direction. This gives the reader a comprehensive understanding of how the field has shaped up in the light of recent state-of-the-art developments, and encourages further dissemination and implementation of novel approaches.

Advances in this research direction would make it easy for anyone to take a picture of a monument or architectural landmark and learn about it in terms of its historical and cultural significance, and also know more about how various design styles have diverged or amalgamated to give rise to these architectural wonders around us. Such developments would not only give an impetus to historical and architectural conservation efforts, but also propel the tourism and education industries.